\documentclass[letterpaper]{article} 
\usepackage{aaai25}  
\usepackage{times}  
\usepackage{helvet}  
\usepackage{courier}  
\usepackage[hyphens]{url}  
\usepackage{graphicx} 
\usepackage{natbib}  
\usepackage{caption} 
\usepackage{booktabs}
\usepackage{algorithm}
\usepackage{algorithmic}
\usepackage{multirow}
\usepackage{xcolor}
\usepackage{tabularx}
\usepackage{amsmath}

\usepackage{newfloat}
\usepackage{listings}
\DeclareCaptionStyle{ruled}{labelfont=normalfont,labelsep=colon,strut=off} 
\floatstyle{ruled}
\newfloat{listing}{tb}{lst}{}
\floatname{listing}{Listing}
\title{J\&H: Evaluating the Robustness of Large Language Models Under Knowledge-Injection Attacks in Legal Domain}

\author{
    Yiran Hu\textsuperscript{\rm 1}\textsuperscript{\rm 2}\equalcontrib,\
    Huanghai Liu\textsuperscript{\rm 1}\equalcontrib,\
    Qingjing Chen\textsuperscript{\rm 1},\
    Ning Zheng\textsuperscript{\rm 1},\
    Chong Wang\textsuperscript{\rm 1},\\
    Yun Liu\textsuperscript{\rm 1},\
    Charles L.A. Clarke\textsuperscript{\rm 2}\thanks{Corresponding Author. charles.clarke@uwaterloo.ca},\
    Weixing Shen\textsuperscript{\rm 1}\thanks{Corresponding Author. wxshen@mail.tsinghua.edu.cn}
}
\affiliations{
    \textsuperscript{\rm 1}School of Law, Tsinghua University\\
    \textsuperscript{\rm 2}David R. Cheriton School of Computer Science, University of Waterloo\\

}

\usepackage{bibentry}

\begin{document}

\maketitle
\begin{abstract}
As the scale and capabilities of Large Language Models (LLMs) increase, their applications in knowledge-intensive fields such as legal domain have garnered widespread attention. However, it remains doubtful whether these LLMs make judgments based on domain knowledge for reasoning. If LLMs base their judgments solely on specific words or patterns, rather than on the underlying logic of the language, the ``LLM-as-judges'' paradigm poses substantial risks in the real-world applications. To address this question, we propose a method of legal knowledge injection attacks for robustness testing, thereby inferring whether LLMs have learned legal knowledge and reasoning logic. In this paper, we propose J\&H: an evaluation framework for detecting the robustness of LLMs under knowledge injection attacks in the legal domain.
The aim of the framework is to explore whether LLMs perform deductive reasoning when accomplishing legal tasks. To further this aim, we have attacked each part of the reasoning logic underlying these tasks (major premise, minor premise, and conclusion generation). 
We have collected mistakes that legal experts might make in judicial decisions in the real world, such as typos, legal synonyms, inaccurate external legal statutes retrieval. However, in real legal practice, legal experts tend to overlook these mistakes and make judgments based on logic. However, when faced with these errors, LLMs are likely to be misled by typographical errors and may not utilize logic in their judgments.
We conducted knowledge injection attacks on existing general and domain-specific LLMs. Current LLMs are not robust against the attacks employed in our experiments. In addition we propose and compare several methods to enhance the knowledge robustness of LLMs. 
\end{abstract}

%
\begin{links}
\link{Code}{https://github.com/THUlawtech/LegalAttack}
\end{links}

%

\section{Introduction}

Large Language Models (LLMs) are increasingly applied to knowledge-intensive fields, such as law~\cite{cui2024chatlaw,huang2023lawyer} and medicine~\cite{Med-PaLM,yang2024zhongjing}. In these fields, LLM agents often act as domain experts~\cite{mei2024llm}, which need to rely on comprehensive domain knowledge and logical reasoning to complete domain-specific tasks~\cite{miao2023selfcheck}. However, the reliability and robustness of LLMs in performing these domain-specific tasks have not been fully verified, thus casting doubt on the trustworthiness of LLM agents when they act as domain experts. In the more general domain, research on the robustness of LLMs~\cite{zhu2023promptbench,wang2023decodingtrust} is based on attacks on synonyms or symbol embeddings of the prompt, but in domain-specific tasks, attacks at the knowledge level also need to be defended against\cite{zhou2024mathattack}. Unlike in the general domain\cite{xu2023llm,ni2023evaluating}, in knowledge-intensive tasks, the introduction of domain knowledge, abstract judgment of facts, and reasoning logic chains are all critical. The model needs to progress through a series of logical reasoning steps to make a final judgment. Unlike the existing reasoning work in math~\cite{zhou2024mathattack} and chemistry~\cite{ouyang2024structured}, our work focuses on reasoning which is both natural language based and requires logical reasoning in natural lanuage. Understanding human language and the logic behind it is more complex than merely learning numbers and operation symbols. 

\textit{Can we trust LLMs in the legal domain?}
LLMs are fragile and small perturbations in the prompt can have a significant impact on their performance. Especially in the knowledge-intensive domains, domain experts will automatically ignore those small errors and changes, making judgments based on logical reasoning. But when LLMs act as domain experts, do they make judgments based on comprehensive domain knowledge? When undertaking complex reasoning, do they make judgments based on a chain of logic within the domain, or do they make judgments based on correlation instead of causal inference?\cite{chen2023knowledge}. 

\begin{figure*}[ht]
  \centering
  \includegraphics[width = 0.7\textwidth]{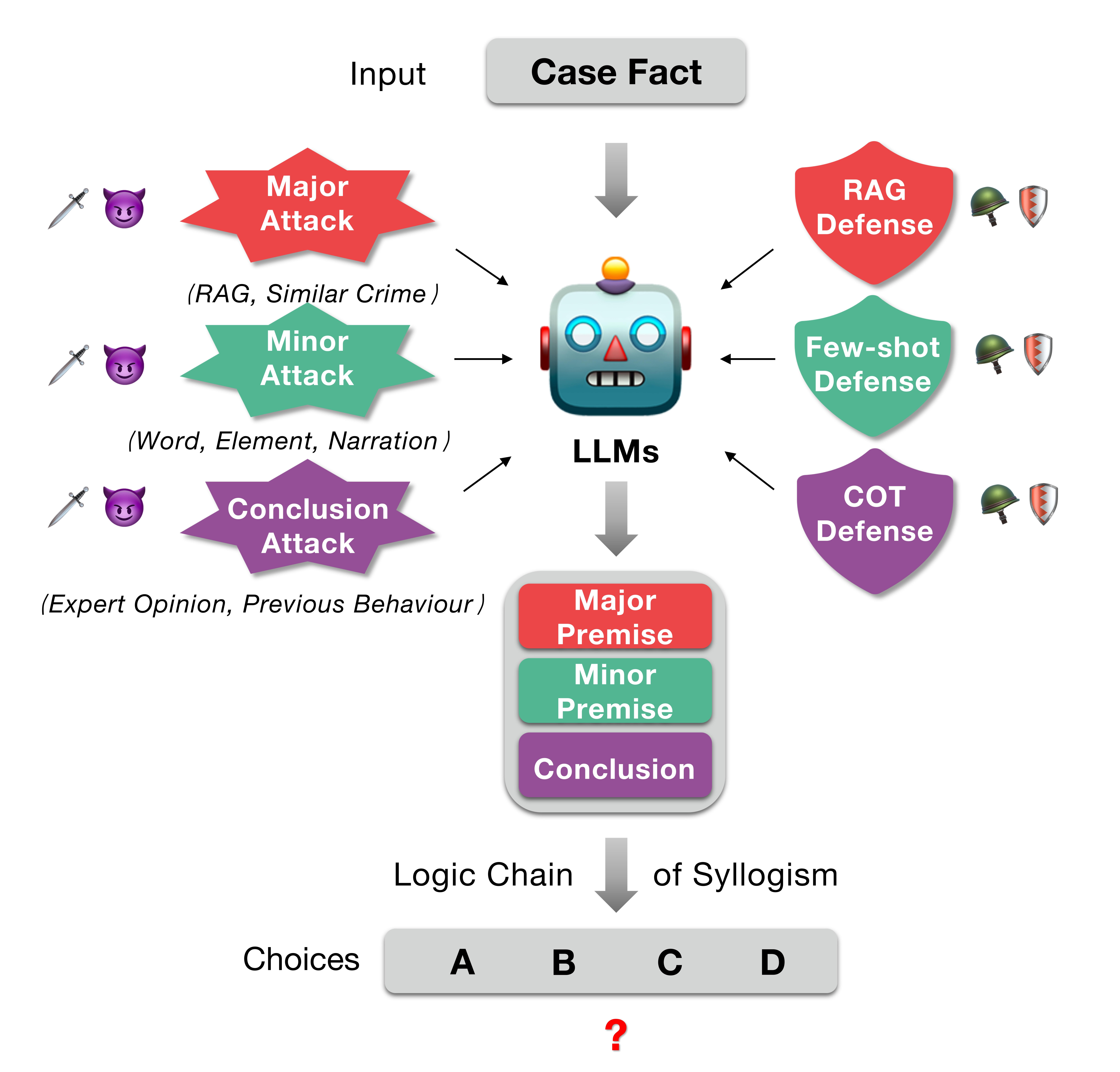} 
  \caption{The Framework of J\&H.} 
  \label{framework} 
\end{figure*}

Based on these considerations, this paper proposes a knowledge attack framework \textbf{J\&H}\footnote{J\&H: Originally taken from an old novel: Jekyll\&Hyde. The protagonist Jekyll who, after being affected by a drug, splits into two personalities: the kind and upright Dr. Jekyll and the demonic Hyde. In this paper, J\&H also stands for Justice \& Hellion. The LLMs could potentially be just, making judgments through domain knowledge and logical inference; but the LLMs could also possibly be a hellion, making judgments without conforming to the logic of the domain. The goal of this paper is to determine whether the LLMs represents Justice or Hellion through knowledge attacks.}  directed at knowledge-intensive domain-specific tasks. In knowledge-intensive fields, judgments require complete reasoning chains and corroboration. In practical reality, domain experts usually employ deductive reasoning to make judgments. Specifically, they adopt the logic chain of syllogism proposed by Aristotle. In this paper, we carry out knowledge attacks according to the logic of syllogism (major premise, minor premise, conclusion). For example, in the medical field, doctors first retrieve possible pathogenesis, and then make disease inferences based on the condition after consultation; in systems based on legal codes, judges first retrieve relevant legal statutes, and then infer the crime based on the legal facts recorded in the trial.


In our work, to test the robustness of LLMs at the level of logical reasoning, we have conducted knowledge attacks J\&H on the three levels of \textbf{``major premise'', ``minor premise'', and ``conclusion generation''}. The framework of J\&H is shown in Figure \ref{framework}. At the major premise level, we perturb the introduced premise. At the factual level of the minor premise, different domains have different factual judgment frameworks\cite{an2022charge,zongyue2023leec}. We have conducted fine-grained annotation J\&H for the legal field. According to the mistakes that judges may make in real-world judgments, we divide the fact-finding part according to the logic of criminal judgment, and manually annotate the domain synonym dictionary for synonym replacement. In the conclusion generation stage, we introduced external disturbance to the conclusion. Throughout the attack process, we ensure that all domain knowledge and facts remain unchanged, so that the attack would not affect the framework of domain experts in reasoning and making judgments. We carry out knowledge injection attacks on each step of the reasoning chain to judge the robustness of LLMs in knowledge-intensive tasks and their reliability at tasks that require logical reasoning.

We conducted attack experiments on existing general-domain LLMs and domain-specific LLMs. The experimental results show that the robustness of these LLMs to knowledge attacks is relatively low. Especially at the conclusion judgment stage, perturbation has a substantial impact on the final judgment. We conducted additional position attacks by inserting noise in the conclusion stage. Results show that inserting noise into the middle part of the prompt will minimally affect the attack effect on the model. In this case, the noise is ``lost in the middle''~\cite{liu2023lost}'

Based on the outcome of our attack experiments, this paper proposes three methods to improve the performance of LLMs under knowledge injection attacks: RAG, COT, and few-shot. However, our experiments show that these three mitigation methods cannot completely and effectively solve the problem of robustness of LLMs against knowledge attacks. This outcome shows that mere modifications at the prompt level cannot completely solve the problem that LLMs cannot use domain knowledge for logical judgment. Therefore, in future research, improvement should be targeted towards the pre-training or fine-tuning process of LLMs.

This paper makes the following contributions:
\begin{enumerate}

\item
\textbf{We propose J\&H: an evaluation framework for evaluating the robustness of LLMs under legal knowledge injection attacks.} In the framework, we use syllogism as the theoretical basis and carry out knowledge attacks on each layer separately.
We conducted fine-grained annotation in the legal field. Dataset annotation includes similar crime name annotation, logical inference annotation, and domain synonym annotation. This annotation framework can be widely applied to other knowledge-intensive fields, and then applied to more domain knowledge attack experiments.

\item
\textbf{We evaluated the existing general domain LLMs and domain-specific LLMs on this benchmark.} We found that current LLMs are susceptible to knowledge injection attacks, lacking robustness under knowledge injection attacks; LLMs cannot use domain knowledge to make correct judgments under the framework of reasoning logic.

\item
\textbf{We propose three ways to enhance robustness: RAG, COT, Few-shot.} Our experiments show that the three methods can enhance the robustness of the model under knowledge injection attacks to a certain extent, but they cannot completely alleviate the problem. This outcome shows that merely through prompting we may not be able to consistently enhance the model's understanding and analysis ability of domain knowledge. Instead, it needs to start at the level of model training and fine-tuning.
\end{enumerate}

\section{Methodology}
\label{method}


\subsection{The J\&H Framework}

The J\&H framework originates from the syllogistic logic of deductive reasoning: \textbf{Major premise - Minor premise - Conclusion}. Logically, the conclusion is derived by applying the major premise to the minor premise. The major premise is a general principle, while the minor premise is a specific statement. As equation \ref{Deductive Reasoning} shows, $\mathcal{A}$ is the major premise, $\mathcal{B}$ is the minor premise and $\mathcal{C}$ is the conclusion.
\begin{equation}
    \mathcal{A} \Rightarrow \mathcal{B}, \mathcal{B} \Rightarrow \mathcal{C} \vdash \mathcal{A} \Rightarrow \mathcal{C}
\label{Deductive Reasoning}
\end{equation}

\begin{figure*}[!h]
  \centering
  \includegraphics[width = 0.85\textwidth]{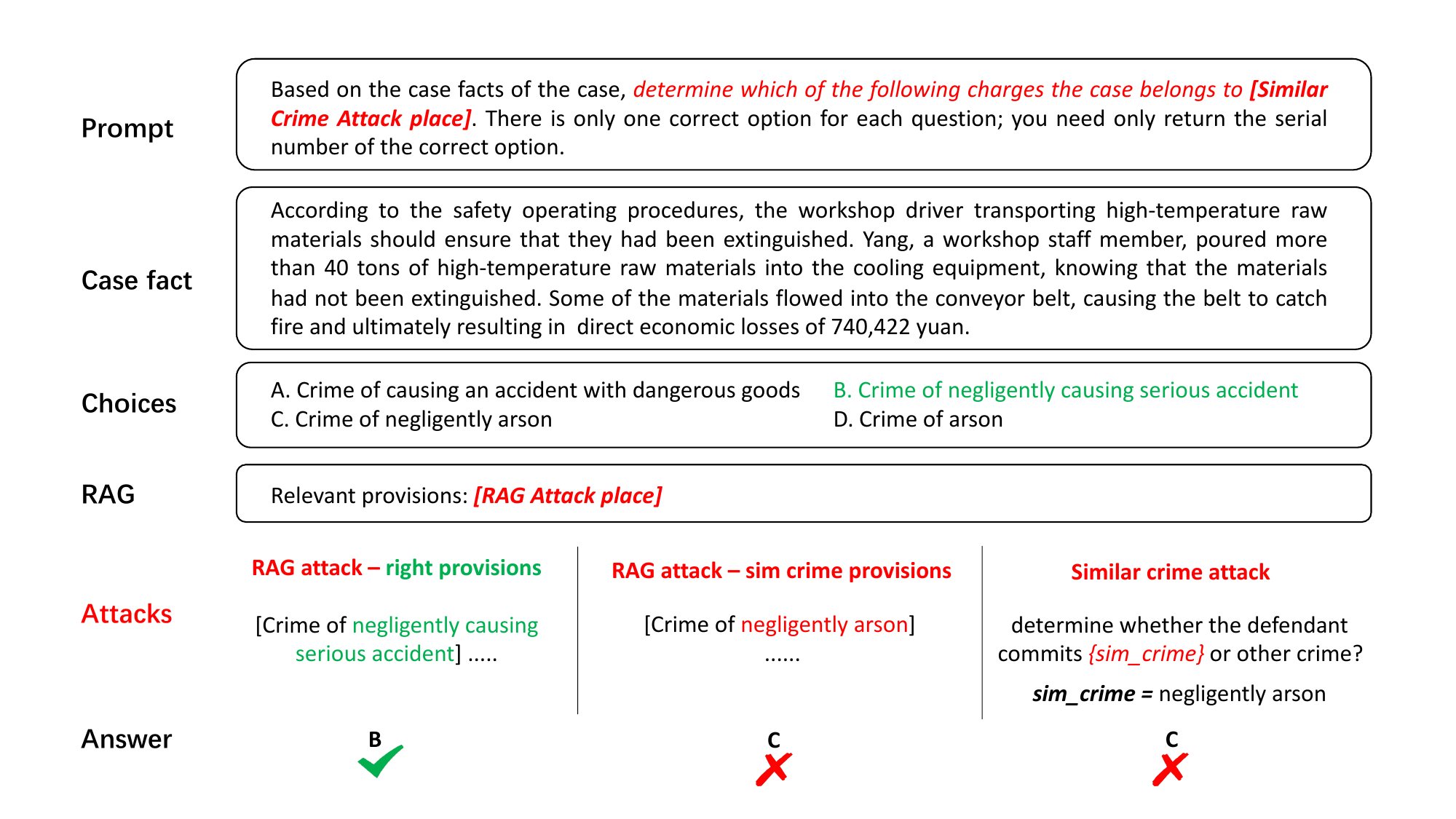} 
  \caption{Illustration of Major Premise Attack.} 
  \label{major} 
\end{figure*}

In knowledge-intensive fields, domain experts conduct rigorous deductions based on syllogistic reasoning to arrive at the final conclusion. Domain experts first seek applicable major premises based on factual circumstances. For example, in the legal field (as shown in Figure\ref{major}) possible crimes such as ``Crime of negligently causing a serious accident'' and ``Crime of arson'' are retrieved based on the fact of ``the ignition of the conveyor belt and the destruction of facilities'', but the difference is that the action of ``Crime of arson'' is negligently causing a fire. Then, real-life facts are transformed into domain knowledge. In the example, the subject is a factory worker, the subject aspect is deliberate, the objective aspect is that he violated the safety of operations by pouring, and the object is public security. Finally the domain knowledge is mapped into the major premise to produce the final conclusion. For the objective ``Crime of arson'' no violation of safety management procedures is necessary and the subject need not be a factory worker, so the crime should not be the ``Crime of arson'', but the ``Crime of negligently causing a serious accident''. When LLMs play the role of domain experts to accomplish tasks, it is not sufficient for them to learn proprietary domain knowledge; they must also understand the associated logical relationships for deduction.



\begin{table*}[]
 \renewcommand\arraystretch{1.3}
\fontsize{7}{8}\selectfont
\resizebox{\textwidth}{!}{
\begin{tabular}
{p{1.5cm}|p{1.5cm}|c|p{6cm}|p{1.9cm}}
\hline
\multicolumn{1}{c|}{\textbf{Prompt}} &
  \multicolumn{4}{p{15cm}}{Based on the facts of the case, determine which of the following crimes the defendant may be guilty of. There is only one correct choice for each question; you need only return the serial number of the correct choice. Case fact: \{\textcolor{red}{\textit{Case fact}}\}. Choices: \{\textcolor{red}{\textit{1 crime + 3 sim\_crimes}}\}. Answer:} \\ \hline
\multicolumn{1}{c|}{\textbf{Attack Level}} &
  \multicolumn{2}{c|}{\textbf{Attack Method}} &
  \multicolumn{1}{c|}{\textbf{Attack Detail}} &
  \textbf{Attack Place} \\ \hline
\multicolumn{1}{c|}{\multirow{3}{*}{Major Premise}} &
  \multicolumn{1}{c|}{\multirow{2}{*}{RAG Attack}} &
  right provisions &
  \multicolumn{1}{c|}{Insert the relevant provisions of the crime for the correct choice} &
  After Choices \\ \cline{3-5} 
\multicolumn{1}{c|}{} &
  \multicolumn{1}{c|}{} &
  sim crime provisions &
  \multicolumn{1}{c|}{Insert the relevant provisions of the crime for the incorrect choice} &
  After Choices \\ \cline{2-5} 
\multicolumn{1}{c|}{} &
  \multicolumn{2}{c|}{Similar Crime Attack} &
  \multicolumn{1}{c|}{Replace with: ``... whether the defendant commits \{\textit{sim\_crime}\} or other crime?"} &
  Prompt, in place \\ \hline
\multicolumn{1}{c|}{\multirow{8}{*}{Minor Premise}} &
  \multicolumn{1}{c|}{\multirow{3}{*}{Word Attack}} &
  common2common &
  \multicolumn{1}{c|}{Select a random word and replace it with a common synonym} &
  Case fact, in place \\ \cline{3-5} 
\multicolumn{1}{c|}{} &
  \multicolumn{1}{c|}{} &
  element2common &
  \multicolumn{1}{c|}{Identify legal four elements and replace them with common synonyms} &
  Case fact, in place \\ \cline{3-5} 
\multicolumn{1}{c|}{} &
  \multicolumn{1}{c|}{} &
  element2element &
  \multicolumn{1}{c|}{Identify legal four elements and replace them with legal synonyms} &
  Case fact, in place \\ \cline{2-5} 
\multicolumn{1}{c|}{} &
  \multicolumn{1}{c|}{\multirow{2}{*}{Element Attack}} &
  factual element &
  \multicolumn{1}{c|}{Insert four elements summarized from the case facts of the corresponding crime} &
  the end of Case fact \\ \cline{3-5} 
\multicolumn{1}{c|}{} &
  \multicolumn{1}{c|}{} &
  provisional element &
  \multicolumn{1}{c|}{Insert four elements summarized from the law of the corresponding crime} &
  the end of Case fact \\ \cline{2-5} 
\multicolumn{1}{c|}{} &
  \multicolumn{1}{c|}{\multirow{3}{*}{Narration Attack}} &
  fine day &
  \multicolumn{1}{c|}{Insert ``It was a fine day."} &
  the end of Case fact \\ \cline{3-5} 
\multicolumn{1}{c|}{} &
  \multicolumn{1}{c|}{} &
  stormy day &
  \multicolumn{1}{c|}{Insert ``It was a stormy day."} &
  the end of Case fact \\ \cline{3-5} 
\multicolumn{1}{c|}{} &
  \multicolumn{1}{c|}{} &
  murder day &
  \multicolumn{1}{c|}{Insert ``It's a dark night, a perfect time for murder."} &
  the end of Case fact \\ \hline
\multicolumn{1}{c|}{\multirow{6}{*}{Conclusion}} &
  \multicolumn{2}{c|}{Previous Behavior Attack} &
  \multicolumn{1}{c|}{Insert ``The perpetrator had previously committed \{\textit{sim\_crime}\}."} &
  the end of Case fact \\ \cline{2-5} 
\multicolumn{1}{c|}{} &
  \multicolumn{1}{p{1.5cm}|}{\multirow{5}{2cm}{Expert Opinion Attack}} &
  pupil &
  \multicolumn{1}{c|}{Insert ``A pupil thought that the perpetrator committed \{\textit{sim\_crime}\}."} &
  the end of Case fact \\ \cline{3-5} 
\multicolumn{1}{c|}{} &
  \multicolumn{1}{c|}{} &
  layperson &
  \multicolumn{1}{c|}{Insert ``A layperson thought that the perpetrator committed \{\textit{sim\_crime}\}."} &
  the end of Case fact \\ \cline{3-5} 
\multicolumn{1}{c|}{} &
  \multicolumn{1}{c|}{} &
  law student &
  \multicolumn{1}{c|}{Insert ``A law student thought that the perpetrator committed \{\textit{sim\_crime}\}."} &
  the end of Case fact \\ \cline{3-5} 
\multicolumn{1}{c|}{} &
  \multicolumn{1}{c|}{} &
  judge &
  \multicolumn{1}{c|}{Insert ``A lawyer thought that the perpetrator committed \{\textit{sim\_crime}\}."} &
  the end of Case fact \\ \cline{3-5} 
\multicolumn{1}{c|}{} &
  \multicolumn{1}{c|}{} &
  lawyer &
  \multicolumn{1}{c|}{Insert ``A judge thought that the perpetrator committed \{\textit{sim\_crime}\}."} &
  the end of Case fact \\ \hline
\end{tabular}}
\caption{Attack methods and details at different levels of J\&H.}
\label{attack-method}
\end{table*}

\begin{figure*}[ht]
  \centering
  \includegraphics[width = 0.9\textwidth]{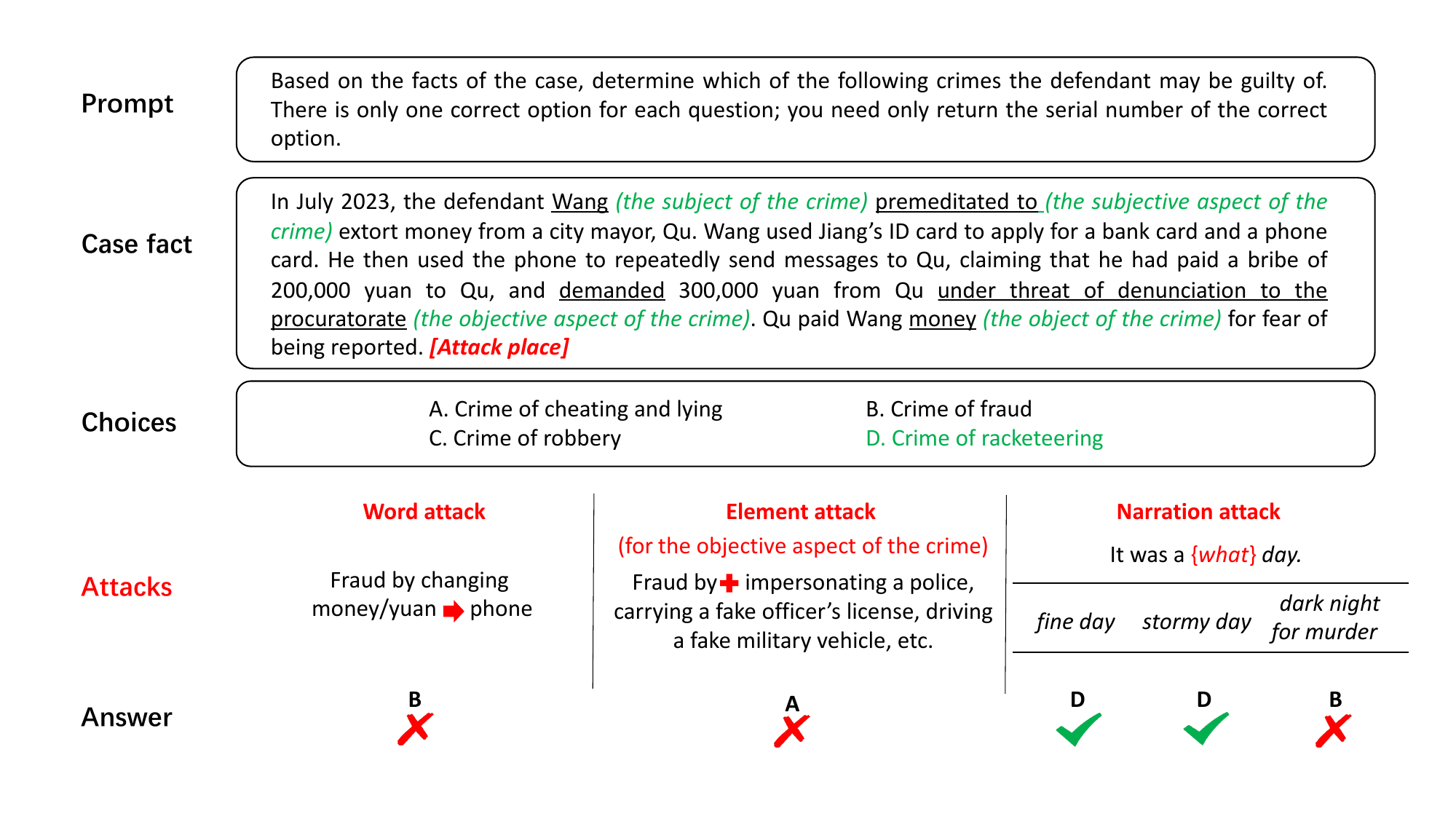} 
   \vspace{-1em}
  \caption{Illustration of Minor Premise Attack.} 
  \label{minor} 
\end{figure*}

In the J\&H framework, in order to evaluate the reliability of LLMs in completing legal tasks, we attack each level of the syllogistic reasoning process in the legal judgment inference.
As shown in Figures \ref{major}, \ref{minor}, and \ref{conclusion-fig}, J\&H has three level attacks. Different levels have different attack methods based on the facts of the case, along with four choices for the conclusion. The choices are generated based on similar crimes related to the correct crime, where similar crimes are those crimes that are easily confused by legal professionals. To create a list of these similar crimes, we invited ten law school graduate students to annotate the cases. In the attack methods of our framework, each choice represents a similar crime. We incorporate these similar crimes into the attack methods, concatenating the attack sentences into the prompt, to check whether the correct choice judged by the LLMs before and after the attack are consistent.

\subsubsection{Attack at the Major Premise Level}
The legal provision plays the role of a major premise in the logical deduction of legal tasks. When legal practitioners solve practical legal problems, they first retrieve the most relevant articles from laws and regulations. These articles also serve as the premise and foundation for all reasoning. In practice, judges would discover through comparison that these facts cannot be applied to the major premise; they would not be misled by the incorrect major premise that shouldn't be used as a reference. Instead, they would determine the correct major premise for judgment of the conclusion. In the J\&H major premise level attack, we insert incorrect major premises as references into the facts to evaluate whether the LLMs can be affected by the incorrect premise.

As Figure \ref{major} shows, we consider two attacks at the major premise level, through the insertion of legal articles and the names of similar crimes.
\begin{enumerate}
    \item \textbf{RAG Attack} We insert the legal articles corresponding to similar crimes as related laws, and note that they can be referred to. The goal of our test is to find whether the model would be misled by the incorrect major premise, and whether it can independently retrieve and apply the correct major premise through the case facts.
    \item \textbf{Similar Crime Attack} We mention similar crimes in the prompt to interfere with the accuracy of the LLMs when inferring major premises.
\end{enumerate}


\subsubsection{Attack at the Minor Premise Level}

As shown in the Table \ref{attack-method}, we consider three types of attacks at the minor premise level: \textbf{Word Attack}, \textbf{Element Attack}, and \textbf{Narration Attack}.  In each type of attack, we introduce a reasoning process with four elements. First identifying the four elements from the case, and then launching targeted attacks on these four elements.

\begin{enumerate}
    \item \textbf{Word Attack} We attack the words in facts of the case by synonym substitution. Based on whether attack words and candidate synonyms belong to common words or legal element words, attack methods are divided into common2common attack, element2common attack, and element2element attack. 
    \item \textbf{Element Attack} We insert adversarial elements from the similar crime at the end of the case facts. The similar elements were divided into factual elements summarized in the facts of the case and provisional elements summarized in the law provisions.
    \item \textbf{Narration Attack} We include environmental descriptions of the events to investigate the effect of subtle semantic changes on the final judgment. According to the depth of background rendering, it is divided into ``fine day'', ``stormy day'', and ``murder day''.
\end{enumerate}

In criminal trials, judges usually make judgments on crimes based on  reasoning about four elements. Analyzing from the constituent elements, every crime has the four elements: 
1) \textbf{the subject of the crime}, which refers to the person who commits the criminal act; 
2) \textbf{the subjective aspect of the crime}, which refers to the psychological state that the subject of the crime has towards the criminal act they commit, and its outcome; 
3) \textbf{the objective aspect of the crime}, which refers to the specific manifestation of the criminal act;  and 4) 
\textbf{the object of the crime}, which refers to the social relationship that is protected by criminal law and violated by the criminal act.

As we need to ensure that the legal facts are not affected as well as the legal logic is preserved before and after the attack, we employed legal experts to annotate the legal synonyms and similar constituent elements of legal elements. The expert annotation contains four parts: ``Similar Crime Annotation'', ``Four Element Annotation'', ``Synonym Word Annotation'', and ``Narration Sentence Annotation''. The annotation details and the examples can be found in supplementary materials.




\subsubsection{Attack at the Conclusion Level}
At the conclusion level, we introduce logical chains that are irrelevant to the reasoning logic, interfering with the original logical mapping relationship between the minor premise and the major premise. We divide  conclusion level attacks into two types: ``Expert Opinion Attack'' and ``Previous Behavior Attack''.

\begin{figure*}[ht]
  \centering
  \includegraphics[width = 0.85\textwidth]{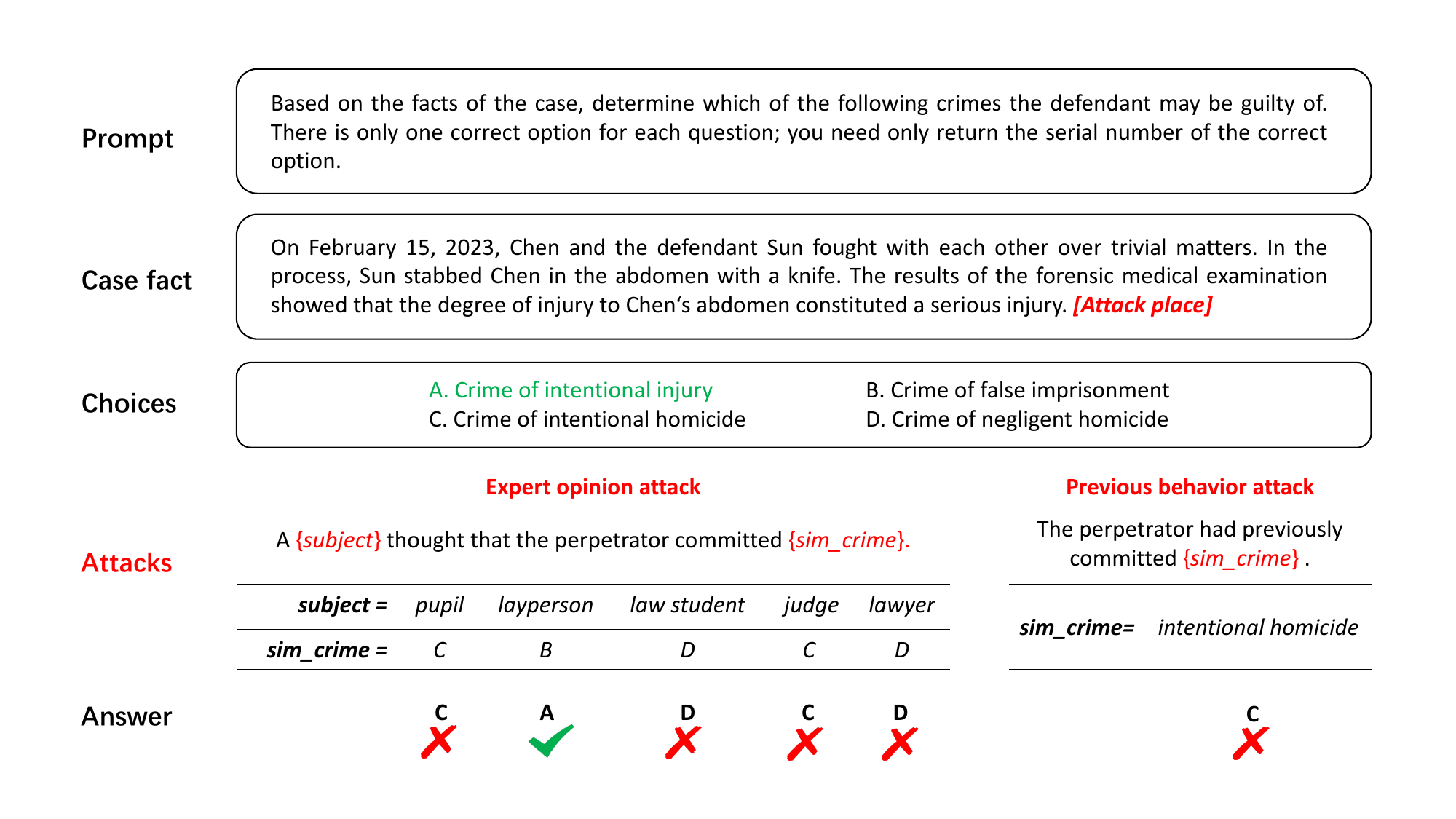} 
   \vspace{-1em}
  \caption{Illustration of Conclusion Attack.} 
  \label{conclusion-fig} 
\end{figure*}

\begin{enumerate}
\item \textbf{Expert Opinion Attack} We insert sentences into the prompt about what crime different identities (from pupils to judges) think the behavior should belong to. The LLMs should ignore the influence of different identities' judgments on the case on its conclusion, and only rely on the facts themselves for logical reasoning. As in the example of Figure \ref{conclusion-fig}, the reasoning judgment of people with legal knowledge can have a negative impact on the LLMs' judgment.
\item \textbf{Previous Behavior Attack} We insert into the prompt the crimes that the perpetrator had previously committed. According to Criminal Law in China, crimes committed by the perpetrator in the past have no impact on the current criminal judgment. The LLMs should not let the logical derivation of the perpetrator's current facts be misled by other logical chains. For example, in the case shown in Figure~\ref{conclusion-fig}, the output result of the large model is affected by the crimes the party has committed before, indicating that the logical chain has been successfully attacked.
\end{enumerate}

\section{Experiments}
\label{experiments}

\subsection{Datasets}
We adopt two legal datasets for our experiments:
\begin{itemize}
\item
\textbf{LEVEN\cite{yao2022leven}} is a large-scale Legal Event Detection dataset, with 8, 116 legal documents and 150, 977 human-annotated event mentions in 108 event types.
\item
\textbf{CAIL2018\cite{xiao2018cail2018}} is the first Chinese legal dataset for judgment prediction. 
\end{itemize}

For our experiments, we use the case facts in these datasets, and the corresponding crime labels are updated according to the latest criminal law in China. The final dataset statistics are shown in Table \ref{data-distribution}. Each question is a multiple-choice question that asks the  LLMs to predict the correct choice based on case facts under different instructions in Table \ref{attack-method}. Each question consisted of one correct crime and three similar crimes, where similar crimes were selected based on annotations from legal experts. All four choices were randomly shuffled.

\begin{table}[h]
\centering
\scalebox{0.9}{
  \begin{tabular}{l|l|l|l|l}
    \toprule
    Dataset & Size  & Charges & Avg length  & Max length\\
    \midrule
    CAIL2018 & 15806  & 184   & 419 & 39586\\
    LEVEN     & 3323 & 61   & 529  & 2476\\
    \bottomrule
  \end{tabular}
}
\caption{Dataset distribution.}
\label{data-distribution}
\end{table}

\begin{table*}[h]
\centering
\scalebox{0.65}{
\begin{tabular}{l|c|cccccc}
\hline
\multicolumn{1}{c|}{\multirow{3}{*}{LEVEN}} &
  \multirow{3}{*}{Original} &
  \multicolumn{6}{c}{Major Premise Level} \\ \cline{3-8} 
\multicolumn{1}{c|}{} &
   &
  \multicolumn{4}{c|}{RAG Attack} &
  \multicolumn{2}{c}{\multirow{2}{*}{Similar Crime Attack}} \\ \cline{3-6}
\multicolumn{1}{c|}{} &       & \multicolumn{2}{c|}{correct provisions} & \multicolumn{2}{c|}{sim crime provisions} & \multicolumn{2}{c}{} \\ \hline
\multicolumn{1}{c|}{} &
  Acc &
  Acc &
  \multicolumn{1}{c|}{PDR} &
  Acc &
  \multicolumn{1}{c|}{PDR} &
  Acc &
  PDR \\
Baichuan2     & 0.777 & 0.84  & \multicolumn{1}{c|}{-8.11\%}  & 0.728    & \multicolumn{1}{c|}{6.31\%}    & 0.653    & 15.96\%   \\
ChatGLM3 &
  0.734 &
  0.834 &
  \multicolumn{1}{c|}{-13.62\%} &
  0.652 &
  \multicolumn{1}{c|}{11.17\%} &
  0.536 &
  26.98\% \\
GPT3.5 &
  0.671 &
  0.798 &
  \multicolumn{1}{c|}{-18.93\%} &
  0.625 &
  \multicolumn{1}{c|}{6.86\%} &
  0.519 &
  22.65\% \\
LLaMA3 &
  0.679 &
  0.834 &
  \multicolumn{1}{c|}{-22.83\%} &
  0.471 &
  \multicolumn{1}{c|}{30.63\%} &
  0.504 &
  25.77\% \\
Farui &
  0.849 &
  0.888 &
  \multicolumn{1}{c|}{-4.59\%} &
  0.824 &
  \multicolumn{1}{c|}{2.94\%} &
  0.758 &
  10.72\% \\ \hline
\end{tabular}
}
\caption{Result of attacks at the Major Premise Level.}
\label{Major-Premise-Level}
\end{table*}

\begin{table*}[!h]
 \renewcommand\arraystretch{1.1}
\resizebox{\textwidth}{!}{
\begin{tabular}{l|c|cccccccccccccccc}
\hline
\multicolumn{1}{c|}{\multirow{3}{*}{LEVEN}} &
  \multirow{3}{*}{Ori} &
  \multicolumn{16}{c}{Minor Premise Level} \\ \cline{3-18} 
\multicolumn{1}{c|}{} &
   &
  \multicolumn{6}{c|}{Word Attack} &
  \multicolumn{4}{c|}{Element Attack} &
  \multicolumn{6}{c}{Narration Attack} \\ \cline{3-18} 
\multicolumn{1}{c|}{} &
   &
  \multicolumn{2}{c|}{com2com} &
  \multicolumn{2}{c|}{ele2com} &
  \multicolumn{2}{c|}{ele2ele} &
  \multicolumn{2}{c|}{factual element} &
  \multicolumn{2}{c|}{provisional element} &
  \multicolumn{2}{c|}{fine day} &
  \multicolumn{2}{c|}{stormy day} &
  \multicolumn{2}{c}{murder day} \\ \hline
 &
  Acc &
  Acc &
  \multicolumn{1}{c|}{PDR} &
  Acc &
  \multicolumn{1}{c|}{PDR} &
  Acc &
  \multicolumn{1}{c|}{PDR} &
  Acc &
  \multicolumn{1}{c|}{PDR} &
  Acc &
  \multicolumn{1}{c|}{PDR} &
  Acc &
  \multicolumn{1}{c|}{PDR} &
  Acc &
  \multicolumn{1}{c|}{PDR} &
  Acc &
  PDR \\
Baichuan2 &
  0.777 &
  0.782 &
  \multicolumn{1}{c|}{-0.64\%} &
  0.773 &
  \multicolumn{1}{c|}{0.51\%} &
  0.722 &
  \multicolumn{1}{c|}{7.08\%} &
  0.681 &
  \multicolumn{1}{c|}{12.36\%} &
  0.534 &
  \multicolumn{1}{c|}{31.27\%} &
  0.773 &
  \multicolumn{1}{c|}{0.51\%} &
  0.775 &
  \multicolumn{1}{c|}{0.26\%} &
  0.765 &
  1.54\% \\
ChatGLM3 &
  0.734 &
  0.738 &
  \multicolumn{1}{c|}{-0.54\%} &
  0.721 &
  \multicolumn{1}{c|}{1.77\%} &
  0.681 &
  \multicolumn{1}{c|}{7.22\%} &
  0.696 &
  \multicolumn{1}{c|}{5.18\%} &
  0.644 &
  \multicolumn{1}{c|}{12.26\%} &
  0.735 &
  \multicolumn{1}{c|}{-0.14\%} &
  0.734 &
  \multicolumn{1}{c|}{0.00\%} &
  0.719 &
  2.04\% \\
GPT3.5 &
  0.671 &
  0.671 &
  \multicolumn{1}{c|}{0.00\%} &
  0.666 &
  \multicolumn{1}{c|}{0.75\%} &
  0.623 &
  \multicolumn{1}{c|}{7.15\%} &
  0.651 &
  \multicolumn{1}{c|}{2.98\%} &
  0.596 &
  \multicolumn{1}{c|}{11.18\%} &
  0.669 &
  \multicolumn{1}{c|}{0.30\%} &
  0.669 &
  \multicolumn{1}{c|}{0.30\%} &
  0.668 &
  0.45\% \\
LLaMA3 &
  0.679 &
  0.688 &
  \multicolumn{1}{c|}{-1.33\%} &
  0.670 &
  \multicolumn{1}{c|}{1.33\%} &
  0.613 &
  \multicolumn{1}{c|}{9.72\%} &
  0.560 &
  \multicolumn{1}{c|}{17.53\%} &
  0.430 &
  \multicolumn{1}{c|}{36.67\%} &
  0.692 &
  \multicolumn{1}{c|}{-1.91\%} &
  0.678 &
  \multicolumn{1}{c|}{0.15\%} &
  0.643 &
  5.30\% \\
Farui &
  0.849 &
  0.847 &
  \multicolumn{1}{c|}{0.24\%} &
  0.845 &
  \multicolumn{1}{c|}{0.47\%} &
  0.803 &
  \multicolumn{1}{c|}{5.42\%} &
  0.807 &
  \multicolumn{1}{c|}{4.95\%} &
  0.746 &
  \multicolumn{1}{c|}{12.13\%} &
  0.846 &
  \multicolumn{1}{c|}{0.35\%} &
  0.846 &
  \multicolumn{1}{c|}{0.35\%} &
  0.825 &
  2.83\% \\ \hline
\end{tabular}
}
\caption{Result of attacks at the Minor Premise Level. `Ori' means the Original results.}
\label{Minor-Premise-Level}
\end{table*}

\subsection{Setup}
We examine the robustness of LLMs in domain-specific tasks on four general LLMs: Azure GPT3.5-turbo\cite{GPT3.5} , Baichuan2-7b-chat\cite{baichuan2}, ChatGLM3-6b\cite{zeng2022glm}, LLaMA3\cite{LLamA3} fine-tuned in Chinese) and one legal-specific LLM: Farui\cite{farui}.







For the open source model, we perform inference on 1 * RTX 4090, and for the closed source model, we call the official API. For truncation of long texts, we sentence-separate the case facts and truncate to the sentence where the case facts + prompt + 100 (space reserved for options, generation, and attacks) $<$ the model's maximum input length.

\subsection{Evaluation Metrics}
Following the approach of Promptbench\cite{zhu2023promptbench}, we use Original Accuracy, Attack Accuracy and Performance Drop Ratio(PDR) as the evaluation metrics. P is the prompt, A is the adversarial attack method, M[x,y] is the evaluation function, which equals to 1 when x=y, and 0 otherwise.

\textbf{Original Accuracy.}
Original Accuracy indicates the accuracy without attack. 
\begin{equation}
Original Acc = \frac{\sum_{(x, y) \in D} \mathcal{M}[f_{\theta}(P,x),y]}{N}
\end{equation}
\textbf{Attack Accuracy.}
Attack Accuracy indicates the accuracy after attack.
\begin{equation}
Acc = \frac{\sum_{(x, y) \in D} \mathcal{M}[f_{\theta}[A(P),x],y]}{N}
\end{equation}
\textbf{Performance Drop Ratio(PDR)}
\begin{equation}
\operatorname{PDR}\left(A, P, f_{\theta}, \mathcal{D}\right)=1-\frac{\sum_{(x ; y) \in \mathcal{D}} \mathcal{M}\left[f_{\theta}([A(P), x]), y\right]}{\sum_{(x ; y) \in \mathcal{D}} \mathcal{M}\left[f_{\theta}([P, x]), y\right]}
\end{equation}
also serves as
\begin{equation}
\operatorname{PDR}=1-\frac{AttackAcc}{OriginalAcc}
\end{equation}

\section{Results and Analytics}

\subsection{Main Results}
We conducted experiments on two datasets using our attack framework. The results from the experiments on LEVEN and CAIL2018 are quite similar. Due to page limit, we report the results from LEVEN in the main body of the paper, and the results from CAIL2018 in the supplementary materials. Experimental results on the Leven dataset can be found in Tables \ref{Major-Premise-Level}, \ref{Minor-Premise-Level} and \ref{Conclusion-level}.

Experimental results show:
\begin{enumerate}
\item
Current LLMs are not robust against the attacks employed in our experiments.
The experimental results show that almost all the adversarial attacks have an impact on the model's output (PDR $>$ 0), and the PDR of many attack methods can exceed 30\%. This result suggests that LLMs are not yet capable of effectively handling domain knowledge when completing domain tasks, nor can they understand the logic of inference in the domain.

\item
Legal attacks are more effective than general attacks. The attack methods that incorporate legal elements are more targeted. For example, in the word attack at the Minor Premise Level, the attack effect of element2common is much worse than that of element2element; for example, in the narration attack under Minor Premise Level, the attack effect of ``fine day'' is worse than that of ``murder day''. This suggests that LLMs cannot accurately judge the difference between legal concepts, so they are easily influenced by legal knowledge attacks.

\item
In dealing with attacks, legal LLMs are more robust than general LLMs. As can be seen from the experimental results, Farui is more robust than other general domain LLMs. This shows that incremental training for the legal domain during the pre-training stage allows LLMs to gain some domain knowledge, but as can be seen, Farui is still not fully robust to our attacks, indicating that LLMs may need to incorporate more domain knowledge through additional fine-tuning.

\item
Among the three levels of attacks, conclusion-level attacks are the most effective. This suggests that LLMs are weak in logical reasoning when handling domain tasks, and their generated conclusions are easily disrupted by conclusion-level adversarial attacks.
\end{enumerate}

\begin{table*}[h!]
 \renewcommand\arraystretch{1.1}
 \resizebox{\textwidth}{!}{
\begin{tabular}{l|c|cccccccccccc}
\hline
\multicolumn{1}{c|}{\multirow{3}{*}{LEVEN}} &
  \multirow{3}{*}{Original} &
  \multicolumn{12}{c}{Conclusion Level} \\ \cline{3-14} 
\multicolumn{1}{c|}{} &
   &
  \multicolumn{2}{c|}{\multirow{2}{*}{Previous Behavior Attack}} &
  \multicolumn{10}{c}{Expert Opinion Attack} \\ \cline{5-14} 
\multicolumn{1}{c|}{} &
   &
  \multicolumn{2}{c|}{} &
  \multicolumn{2}{c|}{pupil} &
  \multicolumn{2}{c|}{layperson} &
  \multicolumn{2}{c|}{law student} &
  \multicolumn{2}{c|}{lawyer} &
  \multicolumn{2}{c}{judge} \\ \hline
\multicolumn{1}{c|}{} &
  Acc &
  Acc &
  \multicolumn{1}{c|}{PDR} &
  Acc &
  \multicolumn{1}{c|}{PDR} &
  Acc &
  \multicolumn{1}{c|}{PDR} &
  Acc &
  \multicolumn{1}{c|}{PDR} &
  Acc &
  \multicolumn{1}{c|}{PDR} &
  Acc &
  PDR \\
Baichuan2 &
  0.777 &
  0.718 &
  \multicolumn{1}{c|}{7.59\%} &
  0.711 &
  \multicolumn{1}{c|}{8.49\%} &
  0.708 &
  \multicolumn{1}{c|}{8.88\%} &
  0.645 &
  \multicolumn{1}{c|}{16.99\%} &
  0.61 &
  \multicolumn{1}{c|}{21.49\%} &
  0.627 &
  19.31\% \\
ChatGLM3 &
  0.734 &
  0.682 &
  \multicolumn{1}{c|}{7.08\%} &
  0.642 &
  \multicolumn{1}{c|}{12.53\%} &
  0.601 &
  \multicolumn{1}{c|}{18.12\%} &
  0.576 &
  \multicolumn{1}{c|}{21.53\%} &
  0.497 &
  \multicolumn{1}{c|}{32.29\%} &
  0.52 &
  29.16\% \\
GPT3.5 &
  0.671 &
  0.66 &
  \multicolumn{1}{c|}{1.64\%} &
  0.547 &
  \multicolumn{1}{c|}{18.48\%} &
  0.549 &
  \multicolumn{1}{c|}{18.18\%} &
  0.514 &
  \multicolumn{1}{c|}{23.40\%} &
  0.528 &
  \multicolumn{1}{c|}{21.31\%} &
  0.478 &
  28.76\% \\
LLaMA3 &
  0.679 &
  0.432 &
  \multicolumn{1}{c|}{36.38\%} &
  0.423 &
  \multicolumn{1}{c|}{37.70\%} &
  0.41 &
  \multicolumn{1}{c|}{39.62\%} &
  0.407 &
  \multicolumn{1}{c|}{40.06\%} &
  0.388 &
  \multicolumn{1}{c|}{42.86\%} &
  0.379 &
  44.18\% \\
Farui &
  0.849 &
  0.806 &
  \multicolumn{1}{c|}{5.06\%} &
  0.743 &
  \multicolumn{1}{c|}{12.49\%} &
  0.748 &
  \multicolumn{1}{c|}{11.90\%} &
  0.676 &
  \multicolumn{1}{c|}{20.38\%} &
  0.66 &
  \multicolumn{1}{c|}{22.26\%} &
  0.555 &
  34.63\% \\ \hline
\end{tabular}
}
\caption{Result of attacks at the Conclusion Level.}
\label{Conclusion-level}
\end{table*}

\subsection{Location Attack}
\begin{figure*}[t]
  \centering
  \includegraphics[width = 1\textwidth]{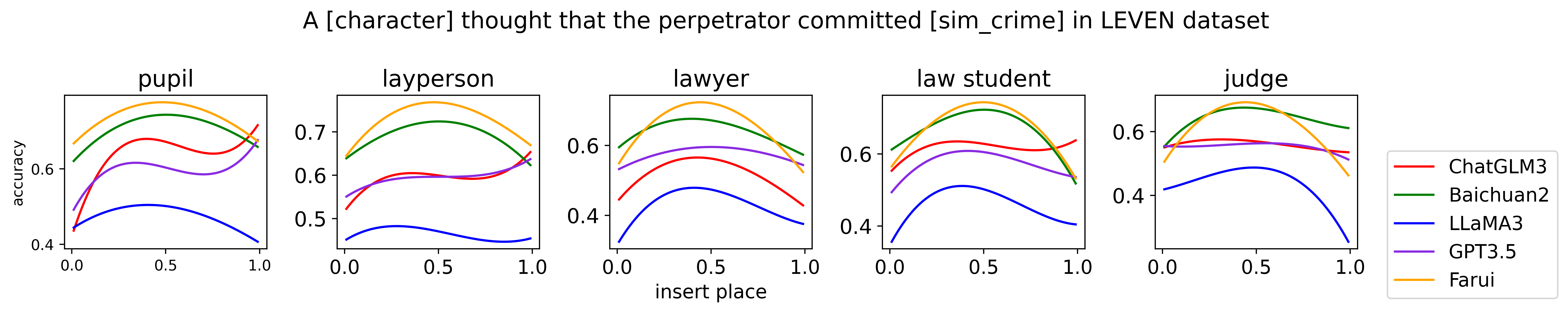} 
  \caption{Location Attack on the LEVEN dataset.} 
  \label{Levenloc} 
\end{figure*}

Given the success of attacks on the conclusion level, we further explored the impact of the attack location~\cite{li2023split}. We conducted location attacks on the expert opinion part of the Conclusion Level on two datasets. We separated the prompt into individual sentences and inserted the expert opinion between the sentences. The final experimental results are shown in Figure~\ref{Levenloc}, with the x-axis representing the insertion position and the y-axis representing Attack Accuracy. As can be seen in the figures, when attacks on the conclusion are placed at the beginning and end, the model is most affected.

\section{Discussion}
\label{discussion}
In this section, we propose three methods to enhance the robustness of the LLMs. Results are shown in Table \ref{discusstion}.

\textbf{RAG}\cite{lewis2020retrieval}. We inserted the legal provision in the criminal law system that is closest to the fact into the prompt and conducted attacks using all methods in the attack framework again. The experimental results show that RAG can improve robustness, but it cannot fully solve this problem.

\textbf{Chain of thought (COT)}\cite{wei2022chain}. We explicitly wrote in the prompt to ``please infer step by step according to the reasoning logic of the four elements of criminal law''. The experimental results show that LLMs do not appear to understand the four elements of criminal law at all, and introducing COT may even make the robustness worse. The model may draw incorrect conclusions through incorrect logic chains.

\textbf{Few-shot.} We inserted two typical cases of the crime and similar crime into the prompt and let the model judge according to the analysis logic of these two cases. The experimental results show that this method also cannot improve the robustness of the model. Large language models appear to be caught in the case details of typical cases and cannot grasp the elements in the case and the logic chain of reasoning.

\begin{table*}[!h]
 \renewcommand\arraystretch{1.1}
 \centering
 \resizebox{0.8\textwidth}{!}{
\begin{tabular}{l|cc|cc|cc|cc}
\hline
\multicolumn{1}{c|}{\multirow{2}{*}{LEVEN}} &
  \multicolumn{2}{c|}{Original} &
  \multicolumn{2}{c|}{RAG} &
  \multicolumn{2}{c|}{COT} &
  \multicolumn{2}{c}{Few-shot} \\
\multicolumn{1}{c|}{} &
  Factual &
  Provisional &
  Factual &
  Provisional &
  Factual &
  Provisional &
  Factual &
  Provisional \\ \hline
Baichuan2 & 12.36\% & 31.27\% & 3.81\% & 15.83\% & 12.58\% & 33.42\% & 0.90\%  & 3.17\%  \\
ChatGLM3  & 5.18\%  & 12.26\% & 6.12\% & 5.40\%  & 5.70\%  & 10.71\% & 0.00\%  & 6.22\%  \\
LLaMA3    & 17.53\% & 36.67\% & 5.88\% & 12.59\% & 17.23\% & 33.72\% & 13.89\% & 33.06\% \\
Farui     & 4.95\%  & 12.13\% & 4.84\% & 11.49\% & 6.37\%  & 15.14\% & 16.26\% & 27.29\% \\ \hline
\end{tabular}
}
\caption{PDR of Element Attack in RAG, COT and Few-shot. After enhancements, the attack is still effective (PDR $>$ 0), but the model is more robust compared to the Original scenario (PDR $<$ Original PDR).}
\label{discusstion}
\end{table*}


\section{Related-Work}
\subsection{General Domain Evaluation}
Existing work\cite{zhu2023promptbench,wang2023decodingtrust,Li_2019,li2020bertattack,morris2020textattack,nie2020adversarial, wang2022adversarial} has made substantial progress on the evaluation of LLMs. AdvGLUE\cite{wang2022adversarial}, DecodingTrust\cite{wang2022adversarial}, PromptBench\cite{zhu2023promptbench} undertake comprehensive benchmarks for evaluating the robustness of LLMs. They focus on the adversarial attacks on input samples as well as the prompts. The attack methods are mainly about the general-domain word level perturbation. Our J\&H is mainly based on knowledge-injection attacks. We propose a knowledge injection attack targeted at LLMs to test their robustness in knowledge-intensive domains. Our attack method is more sophisticated, incorporating not only general semantic interference but also domain knowledge interference annotation, ensuring the accuracy and professionalism of the interference. Furthermore, we introduce logical attacks that conform to the adjudication logic of domain knowledge.
\subsection{Domain-Specific Evaluation}
Previous work~\cite{li2023muser, quan2024econlogicqa, su2024stardchinesestatuteretrieval} has demonstrated that LLMs can be used for the domain-specific tasks, but whether they are reliable when making domain judgments remains unclear. Unlike previous work on domain-specific attacks, such as  MathAttack\cite{zhou2024mathattack}, ChemistryReasoning\cite{ouyang2024structured}, our work depends on the logic underlying language, which can be more complex than numbers and symbols, and it can be widely applied in more knowledge-intensive fields.

\section{Conclusion}
\label{conclusion}
In this paper, we propose a framework of legal knowledge injection attacks for robustness testing for LLMs. We use each part of the deductive reasoning logic to evaluate the models. We evaluate the general-domain and legal-domain LLMs based on the framework. The results show the fragility of prompting LLMs. We also explore several methods to alleviate the issue. We use RAG, COT and Few-shot methods, but the problem still cannot be fully solved. Our experiments show that it is not possible to effectively alleviate the success rate of LLMs being attacked by domain knowledge from the perspective of prompts. Especially in legal tasks, existing LLMs are not reliable and are fragile with respect to prompting. These issues cannot be alleviated by simply improving the prompts. Therefore, in the future it may be necessary to integrate domain knowledge and reasoning chains into the model training process, so that LLMs can be reliable under domain knowledge attacks. J\&H is also available for others working on exploring robust reasoning by LLMs. Researchers can utilize this framework to evaluate the robustness of more LLMs, or apply this framework to more fields related to social life, such as the medical and educational domains.


\appendix

\newpage
\bibliography{aaai25}

\end{document}